\documentclass[3p, preprint, 12pt]{elsarticle}

\usepackage{graphicx}
\usepackage{amssymb}

\usepackage{lineno}
\usepackage{threeparttable}
\usepackage{amsmath}
\usepackage{mathtools}
\usepackage{bm}
\usepackage{subcaption}
\usepackage{booktabs}
\usepackage{hyperref}
\usepackage{multirow}
\usepackage{xcolor}
\usepackage[ruled,vlined]{algorithm2e}
\usepackage{enumitem}
\usepackage{makecell}
\usepackage{xcolor}
\usepackage{import}
\usepackage{graphicx,url}
\usepackage{graphics}
\usepackage{tabularx} 
\usepackage{pdflscape}  
\usepackage{afterpage}
\usepackage{verbatim}

\journal{Cities}

\makeatletter
\def\ps@pprintTitle{%
 \let\@oddhead\@empty
 \let\@evenhead\@empty
 \let\@oddfoot\@empty
 \let\@evenfoot\@empty
}
\makeatother

\begin{document}
\begin{frontmatter}


\title{Generative AI for Urban Design: A Stepwise Approach Integrating Human Expertise with Multimodal Diffusion Models}

 \author[a]{Mingyi He} 
 \author[b]{Yuebing Liang\corref{cor1}}
 \author[c]{Shenhao Wang}
 \author[b]{Yunhan Zheng}
 \author[d]{Qingyi Wang}
 \author[d]{Dingyi Zhuang}
 \author[e]{Li Tian}
 \author[f]{Jinhua Zhao}
 
 \address[a]{Department of Civil and Environmental Engineering, University of California, Berkeley}
 \address[b]{The Singapore-MIT Alliance for Research and Technology}
 \address[c]{Department of Urban and Regional Planning, University of Florida}
 \address[d]{Department of Civil and Environmental Engineering, Massachusetts Institute of Technology}
 \address[e]{Department of Urban Planning, Tsinghua University}
 \address[f]{Department of Urban Studies and Planning, Massachusetts Institute of Technology}
 
 \cortext[cor1]{Corresponding author (ybliang@mit.edu)}

\date{} 

\renewcommand{\thefootnote}{\fnsymbol{footnote}}

\vspace{-.2in}
\begin{abstract}
Urban design is a multifaceted process that demands careful consideration of site-specific constraints and collaboration among diverse professionals and stakeholders. 
The advent of generative artificial intelligence (GenAI) offers transformative potential by improving the efficiency of design generation and facilitating the communication of design ideas. However, most existing approaches are not well integrated with human design workflows. They often follow end-to-end pipelines with limited control, overlooking the iterative nature of real-world design. This study proposes a stepwise generative urban design framework that integrates multimodal diffusion models with human expertise to enable more adaptive and controllable design processes. Instead of generating design outcomes in a single end-to-end process, the framework divides the process into three key stages aligned with established urban design workflows: (1) road network and land use planning, (2) building layout planning, and (3) detailed planning and rendering. At each stage, multimodal diffusion models generate preliminary designs based on textual prompts and image-based constraints, which can then be reviewed and refined by human designers. We design an evaluation framework to assess the fidelity, compliance, and diversity of the generated designs. 
Experiments using data from Chicago and New York City demonstrate that our framework outperforms baseline models and end-to-end approaches across all three dimensions. This study underscores the benefits of multimodal diffusion models and stepwise generation in preserving human control and facilitating iterative refinements, laying the groundwork for human-AI interaction in urban design solutions.

\end{abstract}



\end{frontmatter}

\medskip


\thispagestyle{empty}


\setcounter{footnote}{0}
\renewcommand{\thefootnote}{\arabic{footnote}}
\setcounter{page}{1}

\section{Introduction}
\label{sec:intro}


Urban design encompasses various definitions, one of which - “an activity that adjusts the structural space by producing new spatial organization” - articulates its fundamental nature of spatial manipulation. A more accessible definition, “the process of shaping physical forms of urban areas” reveals the material transformation of urban spaces. Since the 21st century, concepts such as “sustainable place-shaping” and “human-centered environment” have emerged as new trends in urban design \citep{carmona_public_2021}. The significance of well-designed urban design in enhancing residents’ quality of life, fostering economic growth, promoting sustainability, and strengthening social cohesion is indisputable \citep{giles-corti_city_2016, alidoust_master_2022}.

In urban design theory, discussions on the elements and processes of urban design have been ongoing. Among the widely accepted components, “sites, streets, plots, buildings, and open spaces" are considered fundamental \citep{oliveira_urban_2016}. The goal of urban design is to effectively organize these design elements into a coherent urban form. Given the complexity of urban design, real-world projects often adopt a highly hierarchical “top-down" approach as proposed in Conzenian theory \citep{jiang_generative_2024}. This approach involves first laying out road networks and land-use parcels to establish the urban framework, then placing buildings and open spaces within these parcels, and finally integrating detailed design elements such as street furniture and landscaping, providing a useful framework for understanding the organizational structure of urban design \citep{batty_new_2013}.

In practice, the demand for public participation makes urban design a highly iterative and nonlinear process. Each phase of urban design requires collaboration among professionals in urban design, landscape, architecture, and transportation, as well as engagement with stakeholders like regulatory bodies, implementation teams, and public groups \citep{asaad_bridging_2020, quan_urban-gan_2022}. Through continuous negotiation, consensus on the design solution is eventually achieved. For designers, this iterative process of developing ideas can be particularly time-consuming and resource-intensive, as they must respond quickly to various requirements from different stakeholders. However, due to time and resource constraints, only a limited number of design options can be explored, potentially missing opportunities for innovative solutions \citep{jiang_generative_2024}. Additionally, visualized drawings are essential for effectively communicating design ideas in multi-stakeholder contexts \citep{lynch_image_1964}, but their production and rendering further add to the workload \citep{cantrell_digital_2014, ye_masterplangan_2022}.


The rapid advancement of generative artificial intelligence (GenAI) offers promising opportunities to enhance the efficiency of urban design, improving both planning workflows and public participation \citep{park_development_2023,jiang2025urban}. Increasing research has explored GenAI’s ability to generate urban spatial organization patterns. Early studies produced urban images without site-specific constraints, limiting their practical applicability \citep{hartmann_streetgan_2017, albert_modeling_2018}. Subsequent research introduced contextual constraints through image-to-image translation frameworks, enabling spatial alignment between input and output urban images \citep{shen_machine_2020, wu_ganmapper_2022, ye_masterplangan_2022,allen-dumas_generative_2022}. To integrate human input, several studies incorporated design metrics, such as greenery rate and density, as additional model constraints \citep{park_development_2023,
wang_human-instructed_2023, jiang_automated_2024}.

While these studies showcase GenAI’s potential in generating realistic urban diagrams, they often fall short of integrating the expertise of experienced urban planners, limiting their ability to address the complexities of real-world design practices. First, existing approaches offer limited human control. They either neglect human input entirely or reduce it to a small set of predefined design metrics, which lack the flexibility and comprehensiveness needed for meaningful human guidance. Few efforts leverage state-of-the-art multimodal diffusion models with comprehensive control mechanisms \citep{zhang_adding_2023}, leaving significant potential untapped. Second, most methods follow an AI-driven, end-to-end generation process, directly producing design diagrams from site conditions. This approach overlooks the iterative and nonlinear nature of urban design, which involves multiple stages—such as road network and land use planning, building and open space design, and the integration of detailed design elements—before arriving at a final plan. At each stage, urban designers review and refine intermediate designs, coordinating with stakeholders before progressing further. This iterative process enables human oversight and adjustments at key decision points, a critical aspect that end-to-end approaches are unable to accommodate. 

To address these challenges, generative urban design should integrate structured human-AI interaction, providing designers with greater flexibility and control throughout the design process. This study introduces a human-in-the-loop stepwise framework that combines human expertise with multimodal diffusion models to generate urban design diagrams. Aligned with the hierarchical structure of urban design practice, the framework consists of three key phases: (1) road network and land use planning; (2) building layout planning; and (3) detailed planning and rendering. At each phase, textual instructions from human designers are combined with design constraints provided as images to guide the diffusion model in generating corresponding design diagrams. This process enables the iterative creation of diverse and context-specific design solutions, which can then be reviewed and refined by human designers to enhance collaboration among multiple stakeholders. To support open science initiatives, the code is available in the project’s repository at \url{https://github.com/Hemy17/Stepwise_GenerativeUrbanDesign.git}. 
The main contributions of this research are as follows:
\begin{itemize}[noitemsep]
    \item We develop a human-in-the-loop stepwise framework for generative urban design that integrates human expertise with GenAI. Instead of producing a design in a single pass, the framework generates urban design diagrams iteratively, allowing for expert review and refinement at key decision points. This approach ensures a more realistic, adaptable, and interactive design process.

    \item We adapt the state-of-the-art ControlNet diffusion model to generate urban design layouts guided by site constraints and planning guidance. Spatial constraints are encoded as images, and planning guidance is represented as text, enabling controlled and flexible urban design generation that balances automation with human intent.

    \item We design an evaluation framework to assess GenAI's capacity across three key dimensions: visual fidelity, instruction compliance, and design diversity. Experiments in two U.S. cities demonstrate the framework’s superiority over baselines, achieving improved visual quality, higher adherence to human guidance, and greater variation in generated layouts.

\end{itemize}

\section{Literature Review}\label{sec:literature}
\label{sec:theory}

\subsection {Generative AI for Urban Design}\label{sec:literature_urban design}

With the rapid advancement of GenAI, there has been a growing body of research exploring its potential in urban design. Early studies demonstrated the capability of GenAI in synthesizing realistic urban patterns given training examples. \cite{hartmann_streetgan_2017} developed a Generative Adversarial Network (GAN)-based framework, \textit{StreetGAN}, capable of generating street networks that accurately replicate the style of original urban patches.  Similarly, \cite{albert_modeling_2018} proposed a GAN-driven approach to simulate hyper-realistic urban footprints, effectively capturing the complex spatial organization inherent in global urban patterns. \cite{quan_urban-gan_2022} developed Urban-GAN, leveraging a Deep Convolutional GAN (DCGAN) to produce diverse patterns of building layouts that closely resemble real-world urban forms. While these studies highlight GenAI’s potential in replicating realistic urban patterns, they largely focus on unconstrained generation and overlook the integration of specific site conditions or contextual factors, thereby limiting their practical applicability.

To incorporate site conditions in the generation process, subsequent studies have employed image-to-image translation frameworks, enabling the conversion of one urban image into another while preserving spatial alignment. \cite{kang_transferring_2019} adapted image-to-image GANs, including Pix2Pix and CycleGAN, for map style transfer, enabling transformations between different cartographic styles. 
Similarly, \cite{noyman2020deep} developed a streetscape visualization platform that synthesizes realistic street-level images from user-defined viewpoints. 
\cite{ye_masterplangan_2022} introduced a CycleGAN-based framework for smart rendering of urban master plans, converting uncolored design files into fully rendered visualizations. Additionally, studies have leveraged Pix2Pix to generate high-resolution building layouts from input images representing land use characteristics \citep{wu_ganmapper_2022, allen-dumas_generative_2022}. While these studies highlight the effectiveness of image-to-image translation frameworks in preserving site-specific conditions, they overlook the critical role of personalized human guidance in urban design, limiting their adaptability to diverse design scenarios.

To generate diverse design solutions guided by user-specific inputs, recent studies have integrated key design metrics into conditional generative models. 
\cite{wang_deep_2021} and \cite{wang_human-instructed_2023} encoded human-specified metrics, such as greenery rate, into conditional embeddings, which were then combined with environmental data to generate land-use configurations.  \cite{park_development_2023} embedded density metrics—such as floor area ratio (FAR) and building coverage ratio (BCR)—into the RGB channels of input images, framing human-instructed land-use planning as an image-to-image translation task to explore how varying density scenarios affect generated plans. 
For building layout generation, \cite{jiang_building_2023} and \cite{jiang_automated_2024} employed a GAN-based approach that incorporates a conditional vector to encode project-specific requirements, such as target building coverage and height. Despite these efforts, human control over GenAI models remains limited, requiring a more comprehensive and flexible approach. Moreover, most models follow an end-to-end workflow, generating design diagrams directly from site constraints while overlooking the iterative and nonlinear nature of real-world urban design.

\subsection{Image Generation Models}

Existing generative urban design studies predominantly rely on Variational Autoencoders (VAEs) and Generative Adversarial Networks (GANs). VAEs generate new data by learning latent representations and mapping inputs to a probabilistic space for sampling \citep{kingma_auto-encoding_2022, sohn_learning_2015}. In urban studies, VAE-based frameworks have been applied to tasks such as land-use configuration generation \citep{wang_deep_2021, wang_human-instructed_2023}. However, VAE models generally struggle with generating sharp and detailed images \citep{flach_symmetric_2024,cai_enhancing_2024}. GANs, on the other hand, achieve high-quality image generation through adversarial training between a generator and a discriminator \citep{creswell_generative_2018, berthelot_understanding_2018, oring_autoencoder_2020}. In generative urban design, most existing studies rely on GANs, as discussed in the previous section. Despite their success, GANs face challenges of training instability and mode collapse, leading to gradient issues and reduced diversity \citep{saxena_generative_2021}.

In recent years, diffusion models have gained prominence in image synthesis. Their theoretical foundation lies in a two-step process: a forward diffusion phase that incrementally adds noise to images and a reverse diffusion phase that iteratively removes noise to reconstruct high-quality outputs \citep{ho_denoising_2020, dhariwal_diffusion_2021, saharia_palette_2022}. Compared to GANs, diffusion models excel in terms of stability during generation and the superior quality of the produced samples, and thus have become a mainstream choice in image synthesis. In text-to-image generation, diffusion models condition the denoising process on text embeddings, enabling controlled and high-fidelity image synthesis. State-of-the-art models such as DALL·E 2, Imagen, and Stable Diffusion leverage this approach to generate high-resolution, realistic, and diverse images from textual prompts, offering greater flexibility for human control \citep{ramesh_hierarchical_2022, saharia_photorealistic_2022, lugmayr_repaint_2022, li_srdiff_2022, rombach_high-resolution_2022}. 
ControlNet extends Stable Diffusion by integrating external structural constraints, such as edge detection and depth maps, allowing multimodal inputs to guide generation \citep{zhang_adding_2023, wang_-context_2023}. In urban studies, recent research has primarily applied diffusion models to geographical data transformation. \cite{zhuang_advancing_2024} developed a soundscape-to-streetview diffusion model that visualizes urban soundscapes by generating corresponding street-view images, linking auditory and visual perceptions. To address inconsistencies in geographic data quality, \cite{zhou_controlcity_2024} introduced ControlCity, a ControlNet-based framework that synthesizes high-resolution building footprints from multi-source, low-quality data. Despite these advancements, the potential of multimodal diffusion models in controllable urban design generation remains largely unexplored.

\section{Methodology}
\label{sec:method}

\begin{figure}[!ht]
    \centering
    \includegraphics[width=1.0\linewidth]{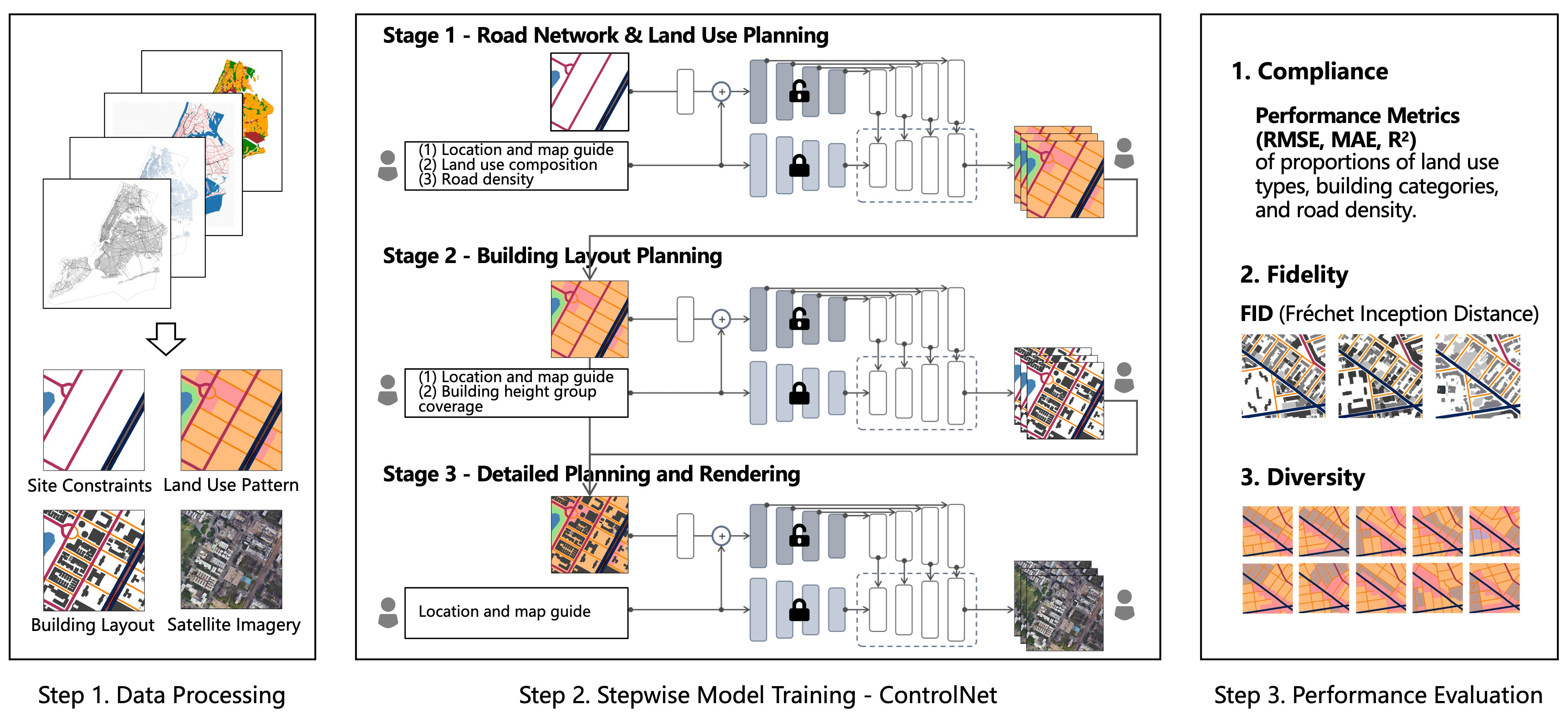}
    \caption{Framework Overview}
    \label{Framework_Overview}
\end{figure}

In this section, we introduce a stepwise generative urban design framework to integrate human expertise and multimodal diffusion models. 
Unlike urban design generation in one go, we decompose the process into three sequential steps: (1) road network and land use planning, (2) building layout planning, and (3) detailed planning and rendering, aligning with established urban design practices. At each stage, human expertise influences the design through prompt input, allowing for control over the generated images, as well as selection, modification, and refinement of the outputs before progressing to the next stage.
The framework starts with data processing to create constraint images, planning guidance, and urban design diagrams for different stages (see Section~\ref{sec:data processing}). Next, we train ControlNet models for stepwise urban design generation, incorporating image constraints and planning guidance at each step (see Section~\ref{sec: stepwise}). Finally, the model’s performance is evaluated across three key dimensions: visual fidelity, instruction compliance, and design diversity (see Section~\ref{sec: evaluation}). An overview of the framework is illustrated in Figure~\ref{Framework_Overview}.

\subsection{Data Processing} \label{sec:data processing}

We collect multi-source data to represent urban spatial layouts. This includes vector data for key urban elements such as roads, railways, water bodies, land use, and building footprints, as well as raster data from satellite imagery, which provides an intuitive visualization of the urban landscape. All data are sourced from open-access repositories to enhance generalizability and reproducibility, with detailed sources introduced in Section~\ref{sec:exp_data}.

\begin{figure}[!ht]
    \centering
    \includegraphics[width=1.0\linewidth]{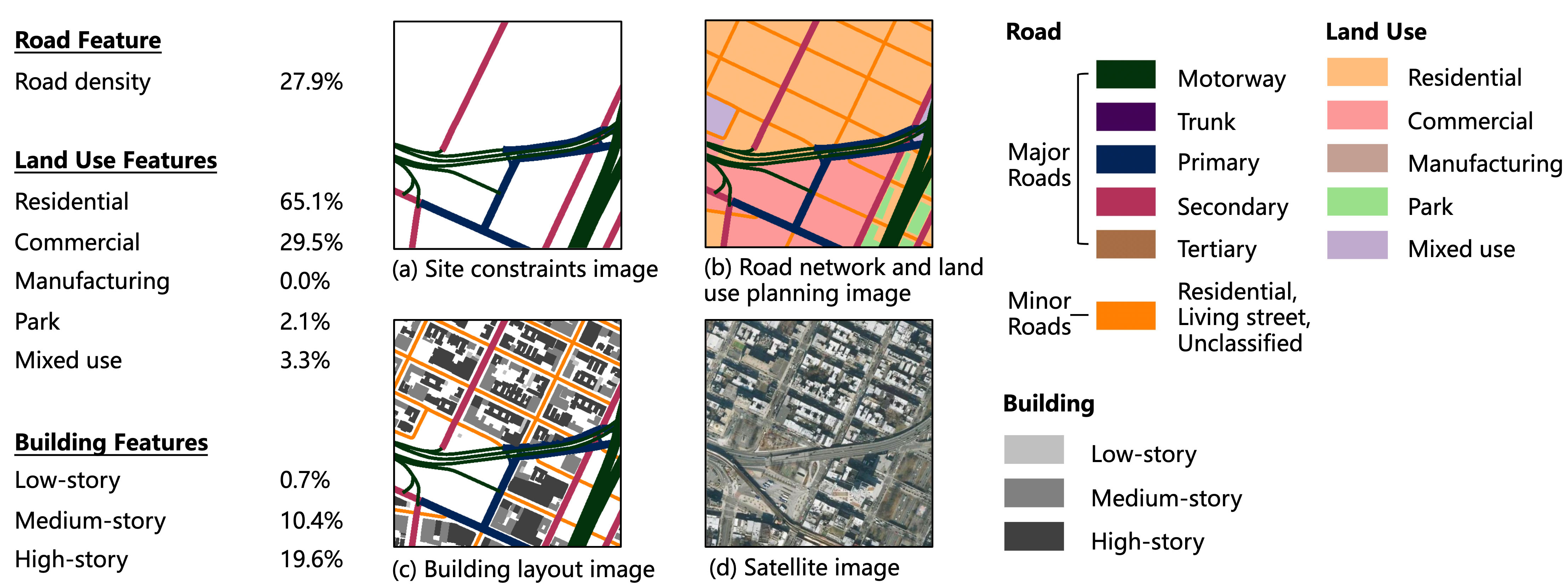}
    \caption{Image Construction and Metric Computation}
    \label{fig:Feature_Extraction}
\end{figure}

To ensure spatial consistency across datasets from different sources and cities, we develop a comprehensive data processing framework. Each city is partitioned into a uniform grid system, with each grid covering an area of 450m × 450m, aligning with the 15-minute city concept \citep{weng_15-minute_2019, rhoads_inclusive_2023}. 
Based on the standardized grid system, we segment urban datasets and construct multiple images that represent site constraints and urban design diagrams across different stages. A visual representation of different images is shown in Figure~\ref{fig:Feature_Extraction}. Specifically, for each grid, we construct four images:
\begin{itemize}[noitemsep]
    \item Site constraints (Figure~\ref{fig:Feature_Extraction}a): Represents initial constraints that urban design must adhere to, such as natural constraints and existing infrastructure. In this study, we include water bodies, railways, and major roads as site constraints, which can be flexibly defined by human designers based on the specific context. 
    \item Road network and land use planning (Figure~\ref{fig:Feature_Extraction}b): Represents the first-stage design decision, where the road network and land use layouts are determined. Based on site constraints, this image adds minor roads and land use layouts. To facilitate cross-city comparison, land use data from different datasets are standardized into five categories: residential, commercial, manufacturing, park, and mixed-use.    
    \item Building layout (Figure~\ref{fig:Feature_Extraction}c): Represents the second-stage design decision, focusing on building placement and height. To ensure consistency across cities, building heights are classified into three categories using the Jenks Natural Breaks method, with each category represented by a distinct grayscale shade.
    \item Satellite image (Figure~\ref{fig:Feature_Extraction}d): Represents the final urban landscape, integrating detailed design elements such as street layouts and rooftop designs. The urban design diagram is rendered into a satellite-style image, offering an intuitive visualization of the urban environment and thereby facilitating public engagement.
\end{itemize}

In addition to image construction, we compute key design metrics as planning guidance. Road features are quantified by road density, defined as the proportion of the area occupied by roads. Land use features are represented by the area proportions of different categories. Building features are captured by the area proportions of open space, low-rise, mid-rise, and high-rise buildings. An example of these metrics are illustrated in Figure~\ref{fig:Feature_Extraction}.

\subsection{Stepwise Model Training} \label{sec: stepwise}

During the training phase, we fine-tune a Stable Diffusion model under the ControlNet framework to generate design images based on textual descriptions and environmental conditions. In this section, we first provide a brief introduction to ControlNet in Section~\ref{sec: model} and then present our stepwise urban design framework in Section~\ref{sec:stepwise_urban_design}.

\subsubsection{Preliminary: ControlNet} \label{sec: model}

Diffusion models are a class of probabilistic generative models grounded in the principles of physical diffusion processes \citep{ho_denoising_2020,dhariwal_diffusion_2021,saharia_palette_2022}. These models consist of a forward diffusion phase, wherein Gaussian noise is incrementally introduced to an input image over a series of steps, progressively degrading it into a highly noisy state. The reverse diffusion phase, facilitated by a neural network, systematically denoises the intermediate representations, ultimately reconstructing the original image with high fidelity.  A key advancement in this field is latent diffusion models (LDMs) \citep{rombach_high-resolution_2022}, which perform denoising in a low-dimensional latent space instead of pixel space, significantly reducing computational costs while maintaining image quality. Building on this foundation, Stable Diffusion\footnote{https://stability.ai/news/stable-diffusion-public-release}, developed by Stability AI, is an open-source LDM optimized for text-to-image generation. It employs a CLIP text encoder to transform text prompts into latent representations, guiding the denoising process to generate corresponding images efficiently.

\begin{figure}[!ht]
    \centering
    \includegraphics[width=0.75\linewidth]{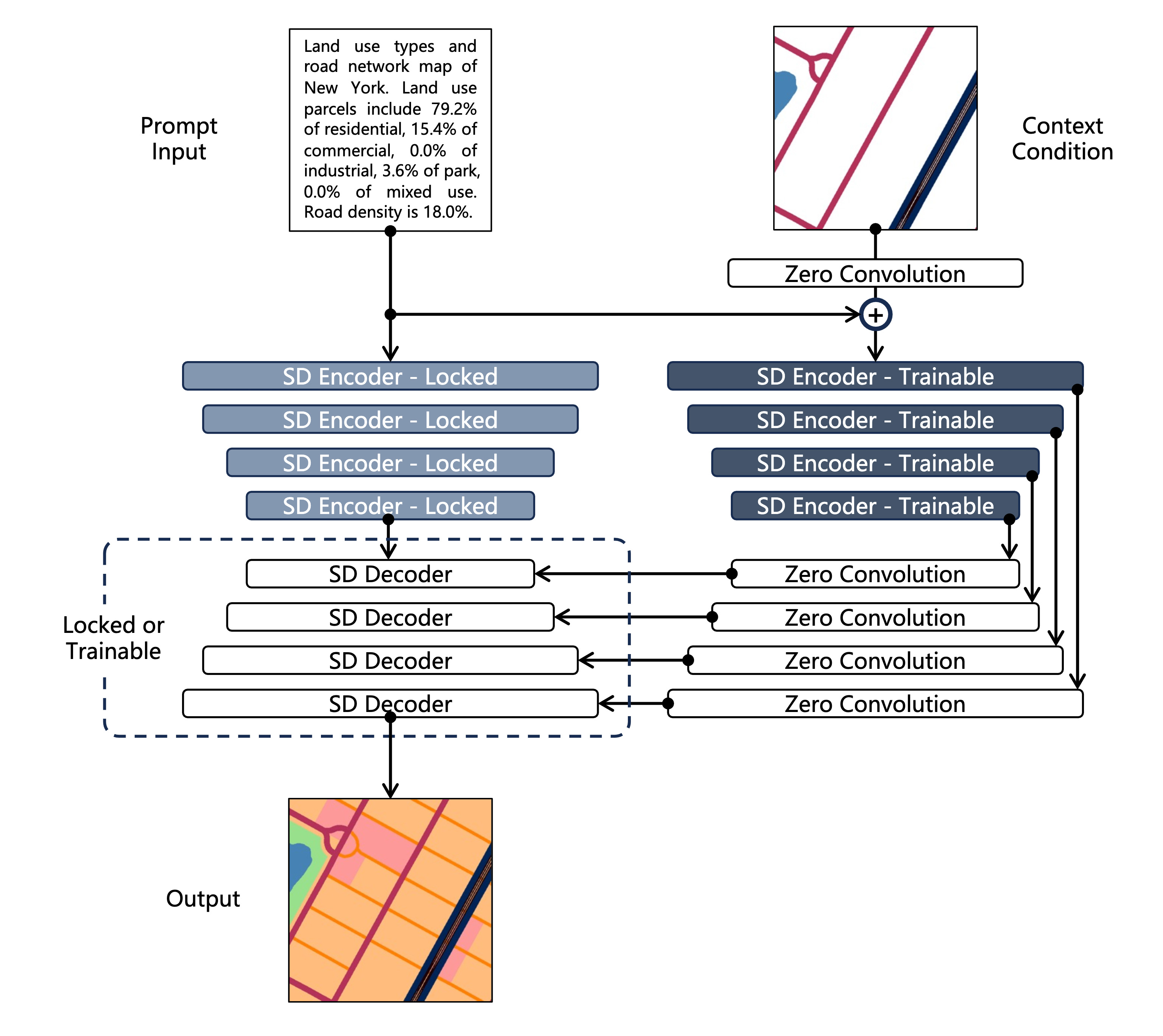}
    \caption{ControlNet Architecture}
    \label{fig:ControlNetArchitecture}
\end{figure}

ControlNet is an advanced neural network architecture that enhances Stable Diffusion by enabling image synthesis guided by additional visual conditions, such as edge maps, depth maps, or human poses \citep{zhang_adding_2023}. Unlike standard Stable Diffusion, which relies solely on text prompts, ControlNet integrates spatial constraints, allowing for more structured and controlled image generation. It achieves this by creating two versions of the neural network blocks: a “locked” copy, which preserves the weights of a pre-trained diffusion model, and a “trainable” copy, which learns to incorporate custom conditioning. During generation, the network’s output is generated as a weighted combination of both versions, guiding the denoising process to align with given structural cues. This dual-network approach ensures greater coherence and fidelity in synthesized images. An overview of its architecture is shown in Figure~\ref{fig:ControlNetArchitecture}.

\subsubsection{ControlNet for Stepwise Urban Design} \label{sec:stepwise_urban_design}

\begin{figure}[!ht]
    \centering
    \includegraphics[width=1.0\linewidth]{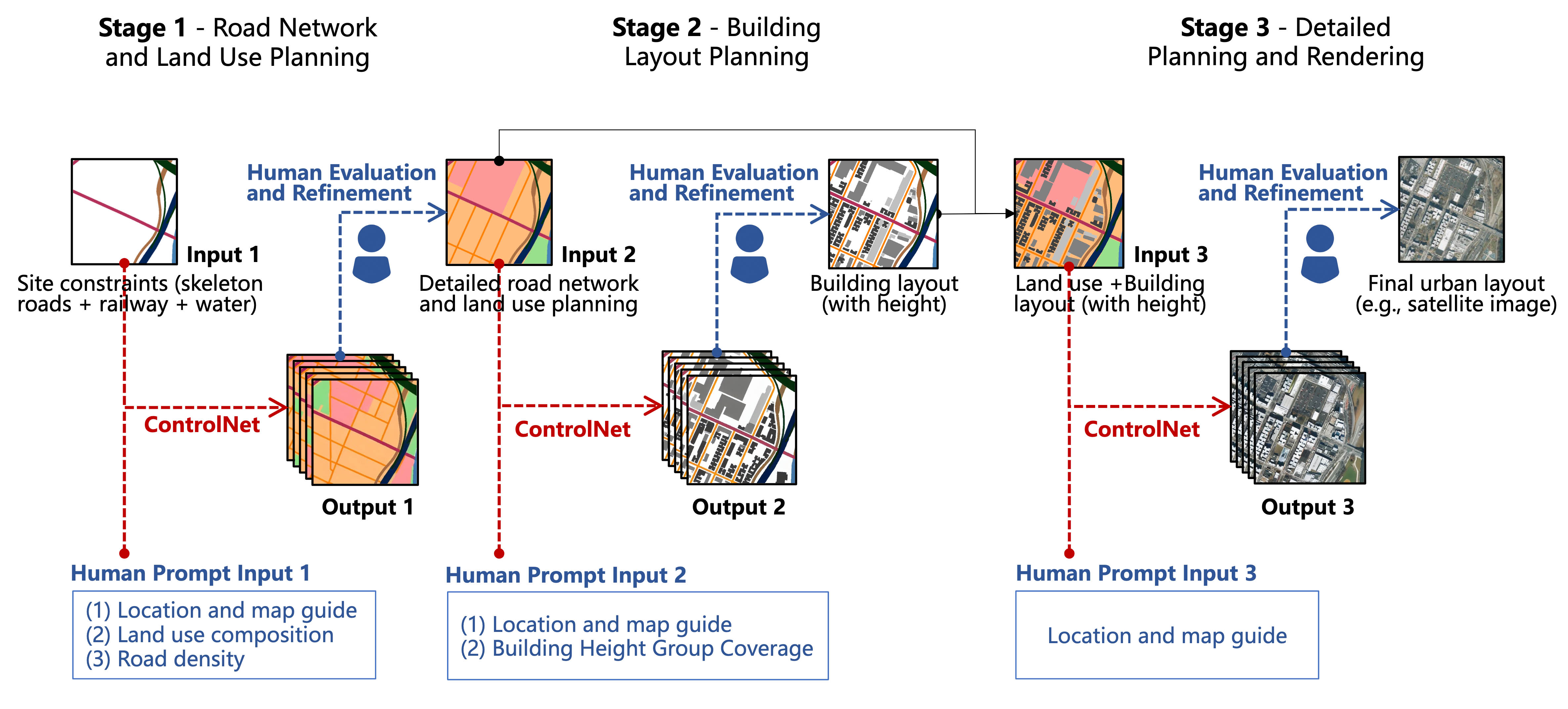}
    \caption{The stepwise urban design framework}
    \label{fig:Stepwise_Approach}
\end{figure}

Based on ControlNet, we propose a human-in-the-loop stepwise urban design framework, as shown in Figure~\ref{fig:Stepwise_Approach}. The framework aligns with established urban design practices and consists of three key stages: (1) road network and land use planning, (2) building layout planning, and (3) detailed planning and rendering. At each stage, a ControlNet model processes context constraints—represented as images—and human control prompts—provided as textual instructions—to generate urban design diagrams that align with the given conditions.

Human expertise is integral to two aspects of the workflow. First, human prompt control directly influences the image generation process. Carefully designed textual prompts serve as structured directives that guide the diffusion model in producing design outputs aligned with specific planning objectives. Second, after each stage of image generation, human expertise is essential for evaluating and refining the generated layouts. From the diverse set of outputs that adhere to the prompt conditions, human designers can select the most suitable design based on conceptual considerations, public engagement outcomes, or value judgments. Additionally, they can modify and refine the selected layouts before feeding them into the next stage of the process. Through this iterative human-AI interaction, the AI-driven rapid generation of diverse design alternatives enhances conceptual exploration efficiency, while human expertise ensures that the final designs remain aligned with professional principles and planning objectives. This cyclical refinement process strikes a balance between computational efficiency and expert-driven control, allowing for a more adaptive and responsive urban design methodology.

At each stage, we carefully design the image constraints and textual prompts to ensure both coherence and flexibility in the design process:

\begin{itemize}[noitemsep]
    \item \textbf{Road Network and Land Use Planning Stage}: As the initial stage of urban design, site-specific constraints, including water bodies, railways, and skeleton road networks, serve as context constraints, with the objective of generating a detailed road network and land use layout. Correspondingly, the textual prompt consists of three components: (1) Location and Map Guide, which defines the city name and the generation task; (2) Land Use Composition, which specifies the percentage distribution of different land use categories to guide land use generation; and (3) Road Density, which indicates the proportion of road area to guide road network generation. An example prompt is provided below: ``\textit{[Location and map guide] Land use types and road network map of New York. [Land use composition] Land use parcels include 79.2\% of residential, 15.4\% of commercial, 0.0\% of industrial, 3.6\% of park, 0.0\% of mixed use. [Road density] Road density is 18.0\%}."
    \item \textbf{Building Layout Planning Stage}: This stage generates building footprints and height distributions. The image constraint is the road network and land use planning generated in the previous stage. The textual prompt consists of two components: (1) Location and Map Guide, which specifies the city name and the generation task; and (2) Building Coverage and Height Distribution, which defines the proportion of open space and the percentage distribution of low-, medium-, and high-story buildings. An example prompt is as follows: ``\textit{[Location and map guide] The Building height gradient map of New York, with shades of gray from light to dark indicating building heights from low to high. [Building height group coverage] The area is composed of 20.50\% low-story buildings, 40.58\% medium-story buildings, 5.64\% high-story buildings, and 33.28\% open space.}"
    \item \textbf{Detailed Planning and Rendering Stage}: This stage renders the urban design diagram into a satellite-style image, providing an intuitive visualization enriched with detailed design elements such as street layouts and landscape features. The image constraint incorporates the planned road network, land use, and building layout from the previous two stages. The textual prompt specifies the city name and generation task, for example: ``\textit{[Location and map guide] Satellite image of a city in New York.}"
\end{itemize}

\subsection{Performance Evaluation} \label{sec: evaluation}

In this study, we develop an evaluation framework to assess GenAI’s capabilities across three key dimensions: visual fidelity, instruction compliance, and design diversity. \textit{Visual fidelity} evaluates whether the generated urban design diagrams look realistic. \textit{Instruction compliance} measures how well the outputs adhere to human-specified prompts. \textit{Design diversity} assesses the model’s ability to generate multiple plausible urban designs under the same constraints. In real-world urban design applications, these three dimensions are all important and should be jointly considered to ensure that GenAI can create realistic, controllable, and diverse design options, thereby supporting human designers in comparison, selection, and refinement:
\begin{itemize}[noitemsep]
    \item \textbf{Visual Fidelity}: Visual fidelity is measured using Fréchet Inception Distance (FID), a widely used metric for evaluating the quality of generated images \citep{ho_denoising_2020}. The FID score quantifies the difference between the distribution of generated images and corresponding real-world images in the feature space of a pretrained InceptionV3 network. Lower FID scores indicate greater fidelity, with generated images more closely resembling real-world urban layouts.
    \item \textbf{Instruction Compliance}: Instruction compliance is evaluated by comparing the generated design diagrams with human-specified design metrics, focusing on the first two stages: (1) road network and land use planning, and (2) building layout planning. In the first stage, we assess the accuracy of land use proportions and road density; in the second, we evaluate open space ratios and building height distributions. Accuracy is measured using root mean square error (RMSE), mean absolute error (MAE), and the coefficient of determination (R²). To account for task complexity in land use generation, we weight each sample by the entropy of its land use distribution, giving more importance to complex cases.
    \item \textbf{Design Diversity}: A third important factor is the diversity of generated images, which ensures that human designers are presented with multiple design alternatives at each stage, thereby supporting comparative selection. To assess design diversity, we employ a qualitative evaluation method. Specifically, we generate multiple outputs using the same input constraints and conduct a visual assessment to examine the extent of variation and the plausibility of each design. 
\end{itemize}

\section{Experiment Settings}\label{sec:exp}

\subsection{Study Area and Data}\label{sec:exp_data}

This study employs two major U.S. cities, New York City (NYC) and Chicago, as case studies.
We utilize multi-source publicly available datasets to ensure the generalizability and reproducibility of the framework. Specifically, road networks, railways, and water bodies are sourced from OpenStreetMap\footnote{https://www.openstreetmap.org/}. Land use data and building footprints, including height information, are acquired from City Open Data Portals\footnote{NYC: https://opendata.cityofnewyork.us/; Chicago: https://data.cityofchicago.org/}. Satellite imagery is obtained from Mapbox\footnote{https://www.mapbox.com/}. The spatial extent of each city is defined by the overlapping coverage of all datasets, ensuring consistency in analysis.

As introduced in Section~\ref{sec:data processing}, we divide each city into a uniform grid system, resulting in 4,049 and 3,209 grids in NYC and Chicago respectively.
To augment the dataset, tiles from both cities are shifted one-third and two-thirds of their original dimensions to the right and downward, respectively, increasing the total sample size ninefold. After augmentation, the datasets comprise 32k and 26k samples for NYC and Chicago, respectively.

\subsection{Model Training and Inference}

To assess GenAI's performance on unseen sites, we carefully design the training and test sets to ensure that all test sites are completely excluded from the training set. Specifically, prior to data augmentation, we randomly select 10\% of the grids as the test set. After augmentation, we exclude any images that overlap with the test set and retain the remaining images as training data. This results in 25,768 training and 317 test images for NYC and 19,738 training and 294 test images for Chicago.

In the ControlNet model, the parameter SD\_locked is set to True by default, keeping certain decoding layers of the stable diffusion model fixed. When set to False, these layers are unlocked, allowing the model to be trained as a whole. While this increases training time and risks performance degradation due to subpar dataset quality, it proves effective for datasets with specific styles or specialized characteristics. In our experiments, training with SD\_locked = False significantly improved the consistency of generated images with input prompts, particularly in road network and building layout stages, enabling better adaptation to unique design requirements. The model was trained for approximately 105 GPU hours on a single NVIDIA L40S GPU (40GB VRAM) with a batch size of 2 and a learning rate of $1 \times 10^{-5}$. 

We compare our framework's performance against several baselines. The first is Pix2Pix, an image-to-image translation framework widely used in generative urban design \citep{kang_transferring_2019,noyman2020deep,wu_ganmapper_2022,park_development_2023}. However, the original Pix2Pix model is limited to image-only inputs. To enable a fair comparison with ControlNet, which incorporates human prompts as control signals, we develop a metric-enhanced Pix2Pix, an extended version that integrates urban design metrics from human instructions as additional conditioning inputs. Additionally, we evaluate ChatGPT-4o, a state-of-the-art industry-level multimodal generation model, and find that it struggles with producing meaningful design diagrams (see \ref{appendix} for details). This highlights the need for domain-specific multimodal GenAI models tailored to urban design.

\section{Results}

The results section consists of five subsections. Sections~\ref{sec: res-fidelity} and \ref{sec: res-compliance} assess visual fidelity and the compliance with human instructions of generated images, demonstrating the superiority of diffusion models over GAN baselines. Section~\ref{sec: res-stepwise} compares stepwise and end-to-end processes in terms of fidelity and compliance, highlighting the advantages of our stepwise framework for urban design.
Section~\ref{sec: res-diversity} visually assesses the diversity characteristics of generated images, emphasizing its suitability to explore multiple design alternatives in generative urban design tasks. Section~\ref{sec: res-transfer} explores cross-city model transferability by testing the model trained in one city on another.

\subsection{Visual Fidelity} \label{sec: res-fidelity}

As introduced in Section~\ref{sec: evaluation}, we use FID to evaluate the visual fidelity of generated images, with lower scores indicating higher fidelity. Table~\ref{tab:FID comparison} presents the quantitative evaluation results, showing that ControlNet consistently outperforms the baseline models across all three generation stages in both cities. The performance gap is particularly pronounced in the first stage, indicating that ControlNet is especially effective in generating realistic road network and land use maps compared to the GAN-based baselines. This advantage likely stems from the superior image synthesis ability of diffusion models, which have been shown to produce higher-fidelity outputs than GANs in various applications \citep{dhariwal_diffusion_2021, saharia_photorealistic_2022}.

\begin{table}[!ht]
    \centering
    \caption{Fidelity performance (FID score) comparison of generated images}
    \begin{tabular}{llccc}
        \hline
        \textbf{Stage} & \textbf{City} & \textbf{ControlNet} & \textbf{Pix2Pix} & \textbf{Metric-enhanced Pix2Pix} \\
        \hline
        \multirow{2}{*}{Stage 1} & NYC & \textbf{80.81} & 185.89 & 224.37 \\
                                 & Chicago  & \textbf{73.97} & 169.50 & 129.55 \\
        \hline
        \multirow{2}{*}{Stage 2} & NYC & \textbf{49.76} & 79.56 & 72.87 \\
                                 & Chicago  & \textbf{52.01} & 64.84 & 61.52 \\
        \hline
        \multirow{2}{*}{Stage 3} & NYC & \textbf{72.52} & 86.21 & 86.21 \\
                                 & Chicago  & \textbf{71.80} & 79.38 & 79.38 \\
        \hline
    \end{tabular}

    \label{tab:FID comparison}
\end{table}

\begin{figure}[!ht]
    \centering
    \includegraphics[width=0.9\linewidth]{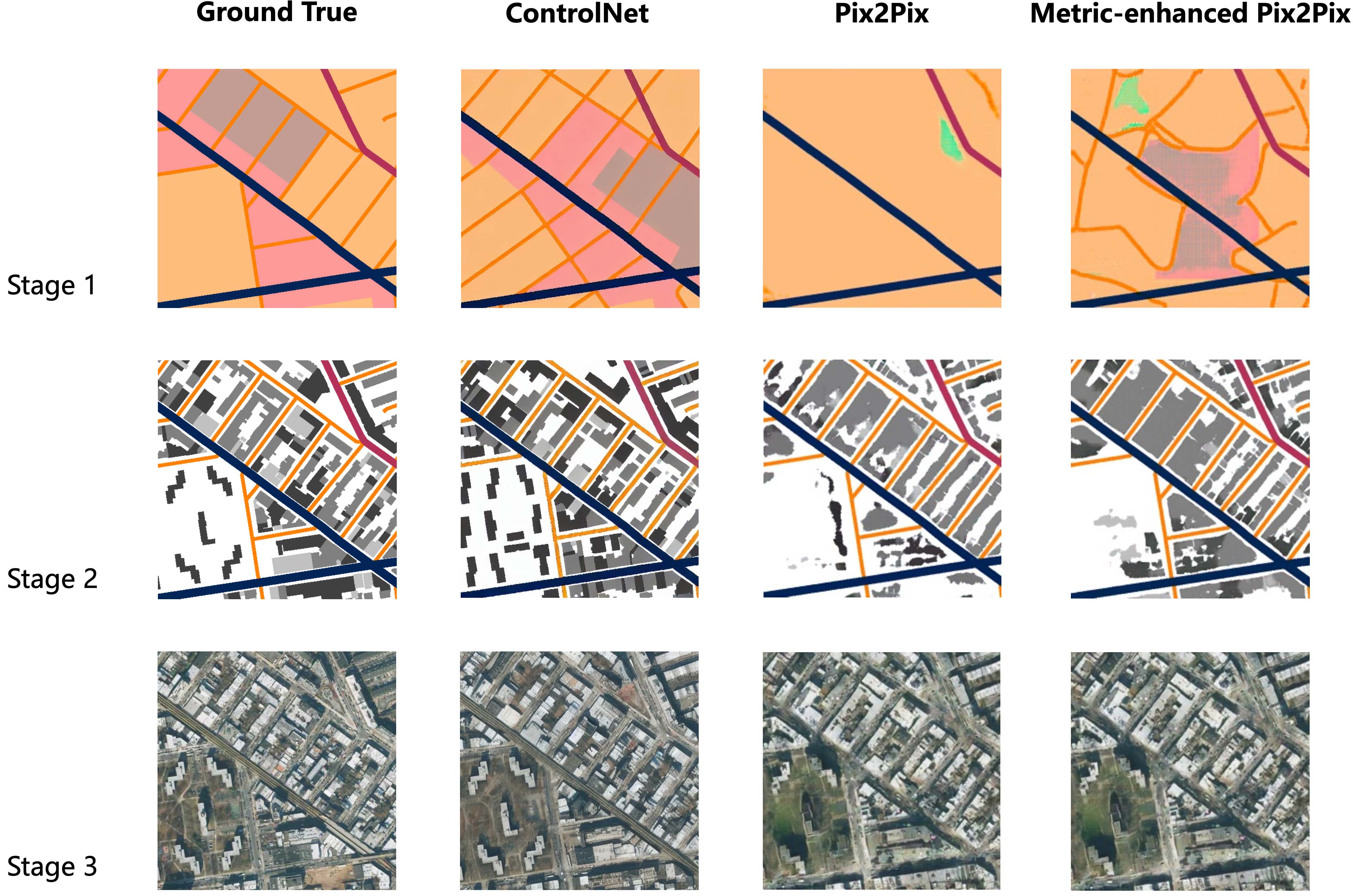}
    \caption{Fidelity comparison of generated images}
    \label{fig:fidelity image}
\end{figure}

To intuitively illustrate differences in visual fidelity, Figure~\ref{fig:fidelity image} presents a comparison of images generated by different models. In the first two stages, ControlNet clearly outperforms the baseline models by producing urban geometries with sharp edges and well-defined boundaries. In the first stage, it generates well-structured road networks that closely resemble real-world layouts, while the baseline models produce distorted or fragmented roads lacking geometric clarity. Furthermore, ControlNet creates land use blocks with regular shapes that are well-integrated with surrounding road networks. In contrast, the baseline models tend to generate irregularly shaped blocks with limited road integration. 
In the second stage, ControlNet produces building footprints that are sharp and clearly defined, typically taking the form of rectangular or orthogonal polygons. In contrast, Pix2Pix-generated footprints exhibit blurry edges and lack clear geometric structure, with only a few recognizable rectangular shapes. In the third stage, while both models generate plausible satellite-style images at first glance, ControlNet produces higher-resolution urban landscapes with well-defined details such as rooftops, street elements, and greenspaces. In contrast, Pix2Pix-generated images exhibit blurriness and lacks sharpness in certain areas, resulting in insufficient detail to reflect realistic urban landscapes.

\subsection{Compliance with Human Instructions} \label{sec: res-compliance}

This section evaluates the performance of our framework in compliance with human instructions. As introduced in Section~\ref{sec: evaluation}, we assess the alignment between the generated images and human-specified design metrics in the first two stages. 

\subsubsection{Road Network and Land Use Planning Stage}\label{sec:compliance:stage1}

In the first stage, we evaluate the model’s compliance with human-specified land use composition and road density. The quantitative results, presented in Table~\ref{tab:stage1}, show that ControlNet largely outperforms the baseline models in aligning with the specified design metrics. Specifically, it achieves high $R^2$ for road density—0.92 in NYC and 0.91 in Chicago—and for land use composition—0.84 in NYC and 0.78 in Chicago. In comparison, Pix2Pix performs poorly in instruction compliance, which is expected since it only accepts image inputs without incorporating human prompts. The metric-enhanced Pix2Pix, which augments the original model by integrating design metrics as additional inputs, performs moderately better but still falls short. It achieves $R^2$ scores of only 0.80 (NYC) and 0.83 (Chicago) for road density, and 0.33 (NYC) and 0.68 (Chicago) for land use composition. These results highlight ControlNet’s effectiveness in generating road layouts and land use maps that more accurately align with human intent.

\begin{table}[!ht]
    \centering
    \renewcommand{\arraystretch}{1.2}
    \caption{Quantitative Evaluation of Instruction Compliance for Stage 1}
    \label{tab:stage1}
    \resizebox{\textwidth}{!}{
    \begin{tabular}{llcccccccccccccc}
        \hline
        \textbf{} & \textbf{City} & \multicolumn{3}{c}{\textbf{ControlNet}} & & \multicolumn{3}{c}{\textbf{Pix2Pix}} & & \multicolumn{3}{c}{\textbf{Metric-enhanced Pix2Pix}} \\
        \textbf{} & & RMSE & MAE & $R^2$ & & RMSE & MAE & $R^2$ & & RMSE & MAE & $R^2$ \\
        \hline
        \multirow{2}{*}{Road Density} & NYC & \textbf{0.02} & \textbf{0.01} & \textbf{0.92} && 0.06 & 0.05 & 0.07 && 0.03 & 0.02 & 0.80  \\
                                 & Chicago  & \textbf{0.02} & \textbf{0.01} & \textbf{0.91} && 0.03 & 0.03 & 0.58 && 0.02 & 0.02 & 0.83 \\
        \hline
        \multirow{2}{*}{Land Use} & NYC & \textbf{0.06} & \textbf{0.04} & \textbf{0.84}  && 0.20 & 0.14 & -0.71 && 0.10 & 0.07 & 0.33 \\
                                 & Chicago  & \textbf{0.06} & \textbf{0.04} & \textbf{0.78}  && 0.16  & 0.12 & -0.12 && 0.08 & 0.06 & 0.68 \\
        \hline
    \end{tabular}
    }
\end{table}

\begin{figure}[!ht]
    \centering
    \includegraphics[width=1.0\linewidth]{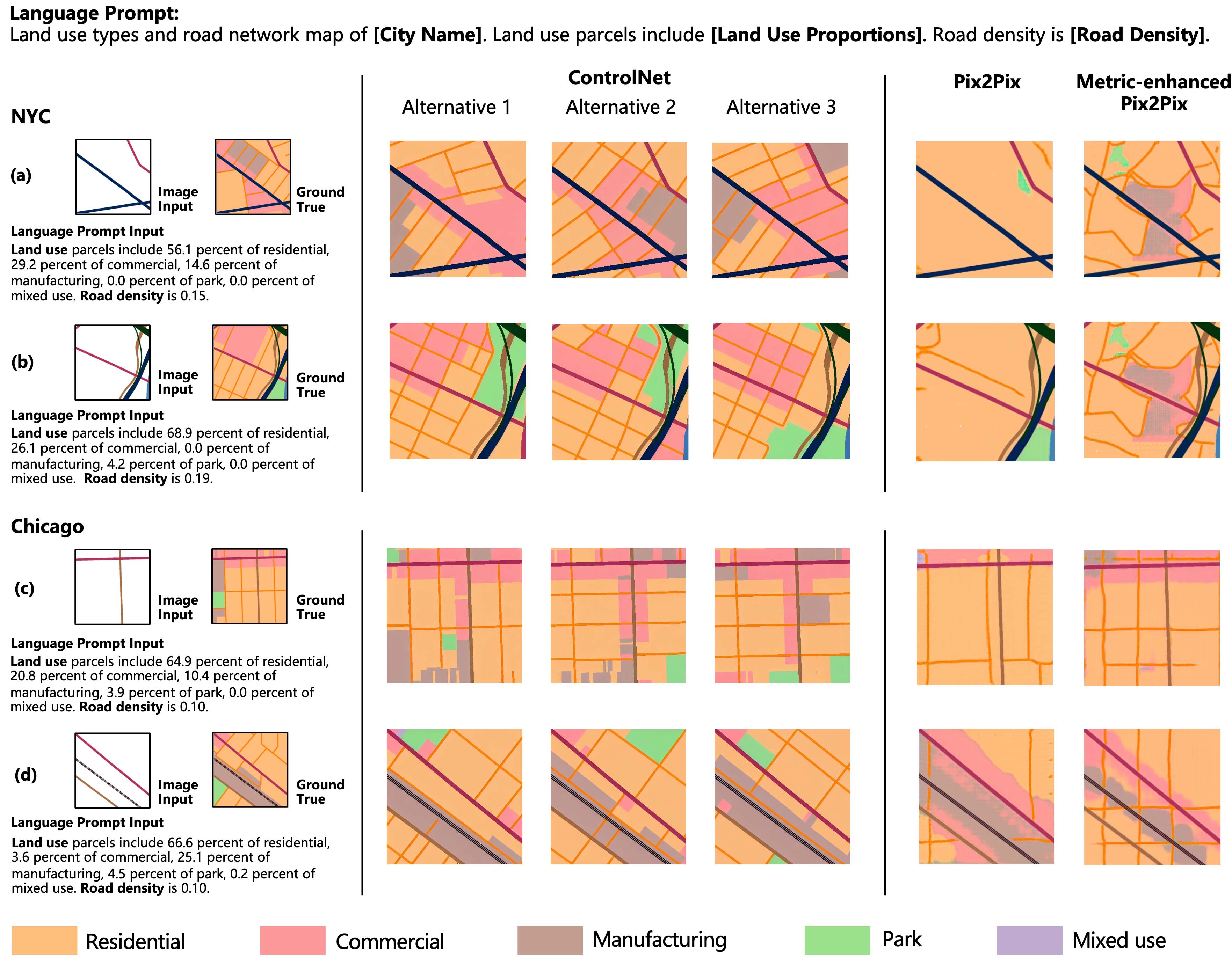}
    \caption{Results of Road Network and Land Use Planning Stage}
    \label{fig:r1}
\end{figure}

Figure~\ref{fig:r1} compares the results generated by ControlNet with those produced by the baseline models under the same constraints. ControlNet effectively captures variations in the instructed road density levels. Figure~\ref{fig:r1}(a)–(d) correspond to target values of 0.15, 0.19, 0.10, and 0.10, respectively. Among them, the generated road network in Figure~\ref{fig:r1}(b) is visibly the densest, followed by (a), with (c) and (d) being the sparsest. Although the street layouts differ from the real-world references, ControlNet successfully produces street blocks that are comparable in size and structure, preserving the intended spatial organization. In contrast, Pix2Pix struggles to generate additional roads beyond the major ones present in the site constraints. While the metric-enhanced Pix2Pix shows slight improvement, it still tends to under-generate road density. Moreover, the resulting road segments are often fragmented with limited connectivity, reducing their practical applicability in urban design contexts. 

Additionally, ControlNet clearly demonstrates a strong ability to interpret and adhere to human-specified land use composition. In Figure~\ref{fig:r1}(a), it successfully generates a mixed-use block containing residential, commercial, and manufacturing areas. While the spatial layout differs from the real-world reference, the generated design adheres closely to the specified land use proportions. Similarly, Figure~\ref{fig:r1}(b) presents a case where the model accurately produces a combination of residential, commercial, and park areas in accordance with the language prompt.

In contrast, Pix2Pix primarily generates a single land use type, failing to represent commercial and manufacturing uses in examples (a) and (b).
While the metric-enhanced Pix2Pix demonstrates improved capacity to create mixed land use compared to the original version, it still over-generates residential land use while under-generating other categories.

Furthermore, ControlNet demonstrates an implicit understanding of spatial relationships between road networks and site constraints. In example (a), it aligns newly generated roads with the diagonal orientation of existing main roads, subdividing the triangular land parcel in an effective way. In example (b), the road network on the northeast side is carefully adjusted to maintain an appropriate spatial distance with the adjacent railway. ControlNet also exhibits awareness of the interaction between land use types and transportation infrastructure. For instance, in examples (a) and (c), commercial areas are primarily located along high-order roads, while residential zones are placed adjacent to lower-order streets—an arrangement consistent with real-world planning practices. In example (b), the park is strategically placed around the railway area, reflecting a common design approach that uses green space as a buffer between transportation infrastructure and residential zones.

\subsubsection{Building Layout Planning Stage}\label{sec:compliance:stage2}

In the second stage, the model generates building footprints and height distributions based on the road network and land use map, along with human-specified design instructions concerning building heights and open space proportions. 
The quantitative results, shown in Table~\ref{tab:stage2}, demonstrate that ControlNet outperforms the baseline models in building‐height distribution accuracy, while achieving comparable open‐space compliance.
Specifically, for building height distribution, ControlNet achieves $R^2$ scores of 0.87 in NYC and 0.81 in Chicago, compared to 0.82 and 0.71 for the metric-enhanced Pix2Pix model. 
In terms of open space compliance, ControlNet outperforms Pix2Pix in Chicago, achieving an $R^2$ of 0.90 compared to 0.84. In New York City, the two models perform similarly, with ControlNet and Pix2Pix reaching $R^2$ values of 0.91 and 0.92, respectively, both demonstrating strong quantitative alignment.

\begin{table}[!ht]
    \centering
    \renewcommand{\arraystretch}{1.2}
    \caption{Quantitative Evaluation of Instruction Compliance for Stage 2}
    \label{tab:stage2}
    \resizebox{\textwidth}{!}{
    \begin{tabular}{llcccccccccccccc}
        \hline
        \textbf{} & \textbf{City} & \multicolumn{3}{c}{\textbf{ControlNet}} & & \multicolumn{3}{c}{\textbf{Pix2Pix}} & & \multicolumn{3}{c}{\textbf{Metric-enhanced Pix2Pix}} \\
        \textbf{} & & RMSE & MAE & $R^2$ & & RMSE & MAE & $R^2$ & & RMSE & MAE & $R^2$ \\
        \hline
        \multirow{2}{*}{Building Height} & NYC & \textbf{0.03} & \textbf{0.02} & \textbf{0.87} && 0.06 & 0.03 & 0.62 && 0.04 & 0.03 & 0.82 \\
                                 & Chicago  & \textbf{0.04} & \textbf{0.02} & \textbf{0.81}     && 0.06 & 0.04 & 0.49 & & 0.05 & 0.03 & 0.71 \\
        \hline
        \multirow{2}{*}{Open Space} & NYC & \textbf{0.05} & \textbf{0.03}  & 0.91  && 0.07 & 0.05 & 0.81 && \textbf{0.05} & \textbf{0.03} & \textbf{0.92} \\ 
                                 & Chicago  & \textbf{0.05} & \textbf{0.04}  & \textbf{0.90}  && 0.08 & 0.05 & 0.77 && 0.06 & \textbf{0.04} & 0.84 \\
        \hline
    \end{tabular}
    }
\end{table}

\begin{figure}[!ht]
    \centering
    \includegraphics[width=0.95\linewidth]{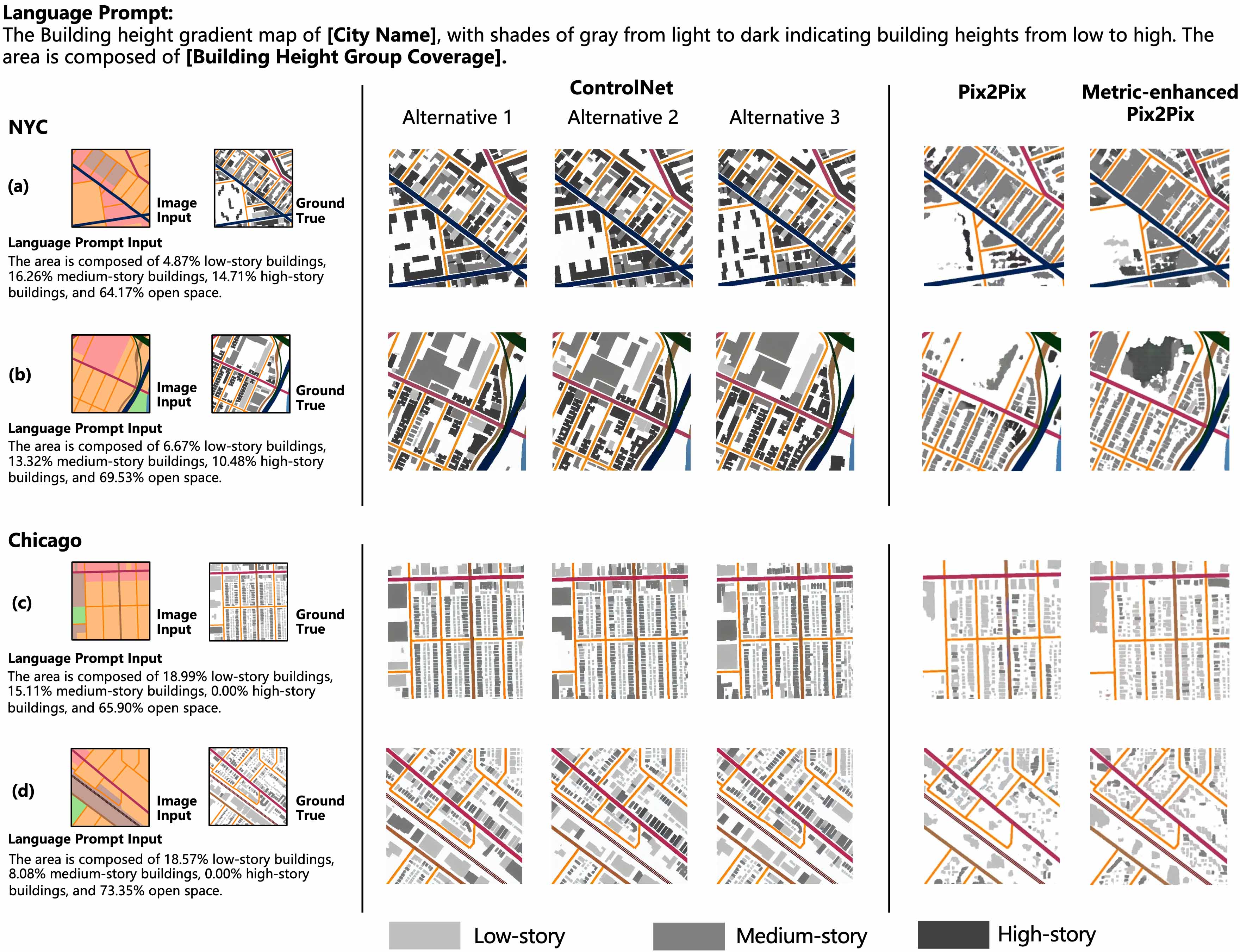}
    \caption{Results of Building Layout Planning Stage}
    \label{fig:r2}
\end{figure}

Figure~\ref{fig:r2} presents examples generated during the building layout planning stage, where gray blocks represent building footprints and darker shades indicate taller buildings. These examples highlight ControlNet’s capacity to reflect variations in building height distributions based on planning guidance. In examples (a) and (b), the input specifies a predominance of medium- and high-story buildings, while examples (c) and (d) call for layouts primarily composed of low-story buildings, followed by medium-story structures. This variation is accurately captured in the generated outputs: building blocks in (a) and (b) appear visibly darker than those in (c) and (d), consistent with the specified height distributions.
In contrast, Pix2Pix exhibits notable deviations from the intended design. In examples (a) and (b), the model significantly under-generates the proportion of high-story buildings, while in example (d), it considerably over-generates the share of low-rise buildings, resulting in an excessively low building density.

Furthermore, ControlNet exhibits a profound understanding of the relationship between road networks and building layouts. In ControlNet-generated images, buildings are consistently aligned along streets, forming a continuous street interface. Specially, in example (c), the generated images successfully reflect real-world patterns, where taller buildings are positioned along streets while shorter structures are placed within parcel interiors. In contrast, the baseline models struggle to maintain street continuity, often producing fragmented building interfaces, particularly evident in examples (b) and (d). Additionally, ControlNet effectively captures the relationship between land use and building morphology.
For instance, in example (b), it generates large building blocks for the square-shaped commercial area and smaller structures for the surrounding residential zone—an arrangement aligned with common design practices. 
In example (d), ControlNet generates larger and taller buildings in manufacturing zones, consistent with the spatial characteristics of real-world industrial facilities. In comparison, the baseline models generate uniformly small and scattered buildings across different land use areas, demonstrating limited sensitivity to the distinct morphological patterns associated with each land use type.

\subsection{Stepwise versus End-to-End Framework} \label{sec: res-stepwise} 

To evaluate the effectiveness of our stepwise framework, we implement an end-to-end variant that generates the urban design diagram in a single step from site constraints. For a fair and quantitative comparison of instruction compliance, this variant is designed to correspond to the first two stages of our framework. Instead of producing intermediate land use and road network outputs, it employs a ControlNet model to directly generate combined road network and building layout diagrams from the initial site constraints. The model is conditioned on textual prompts specifying road density, land use composition, building height distribution, and open space ratio, as defined in the first two stages of the stepwise process.

Table~\ref{tab:stepwise_vs_end2end} compares the stepwise and end-to-end frameworks in terms of visual fidelity and instruction compliance. The stepwise approach performs significantly better in image realism, with an FID score of 49.76, compared to 74.70 for the end-to-end framework. In terms of instruction compliance, the end-to-end framework achieves $R^2$ scores of 0.44 for road density, 0.78 for building height, and 0.48 for open space, whereas the stepwise framework scores 0.92, 0.87 and 0.91, respectively, demonstrating stronger adherence to human instructions. These results suggest that introducing intermediate decision points enables the stepwise approach to generate more realistic urban design diagrams that better align with human instructions. Additionally, from the perspective of urban design practices, the stepwise framework aligns better with real-world planning needs, as it allows for better human intervention at each stage. Therefore, from the perspectives of output quality, alignment with instructions, and better human control, the stepwise approach proves superior to end-to-end.

\begin{table}[!ht]
\centering
\renewcommand{\arraystretch}{1.2}
\caption{Comparison of Stepwise and End-to-End Framework}
\label{tab:stepwise_vs_end2end}
\begin{tabular}{lccccccc}
\hline
\multicolumn{1}{c}{} & \multicolumn{3}{c}{\textbf{Stepwise}} && \multicolumn{3}{c}{\textbf{End-to-end}} \\ \hline
FID & \multicolumn{3}{c}{\textbf{49.76}} && \multicolumn{3}{c}{74.70} \\ \hline
\textbf{} & RMSE & MAE & R\textsuperscript{2} && RMSE & MAE & R\textsuperscript{2} \\ \hline
Road density & \textbf{0.02} & \textbf{0.01} & \textbf{0.92} && 0.05 & 0.04 & 0.44 \\
Building Height  & \textbf{0.03} & \textbf{0.02} & \textbf{0.87} && 0.04 & 0.02 & 0.78 \\ 
Open Space & \textbf{0.05} & \textbf{0.03} & \textbf{0.91}  && 0.12 & 0.09 & 0.48 \\ 
\hline
\end{tabular}
\end{table}

\begin{figure}[!ht]
    \centering
    \includegraphics[width=0.9\linewidth]{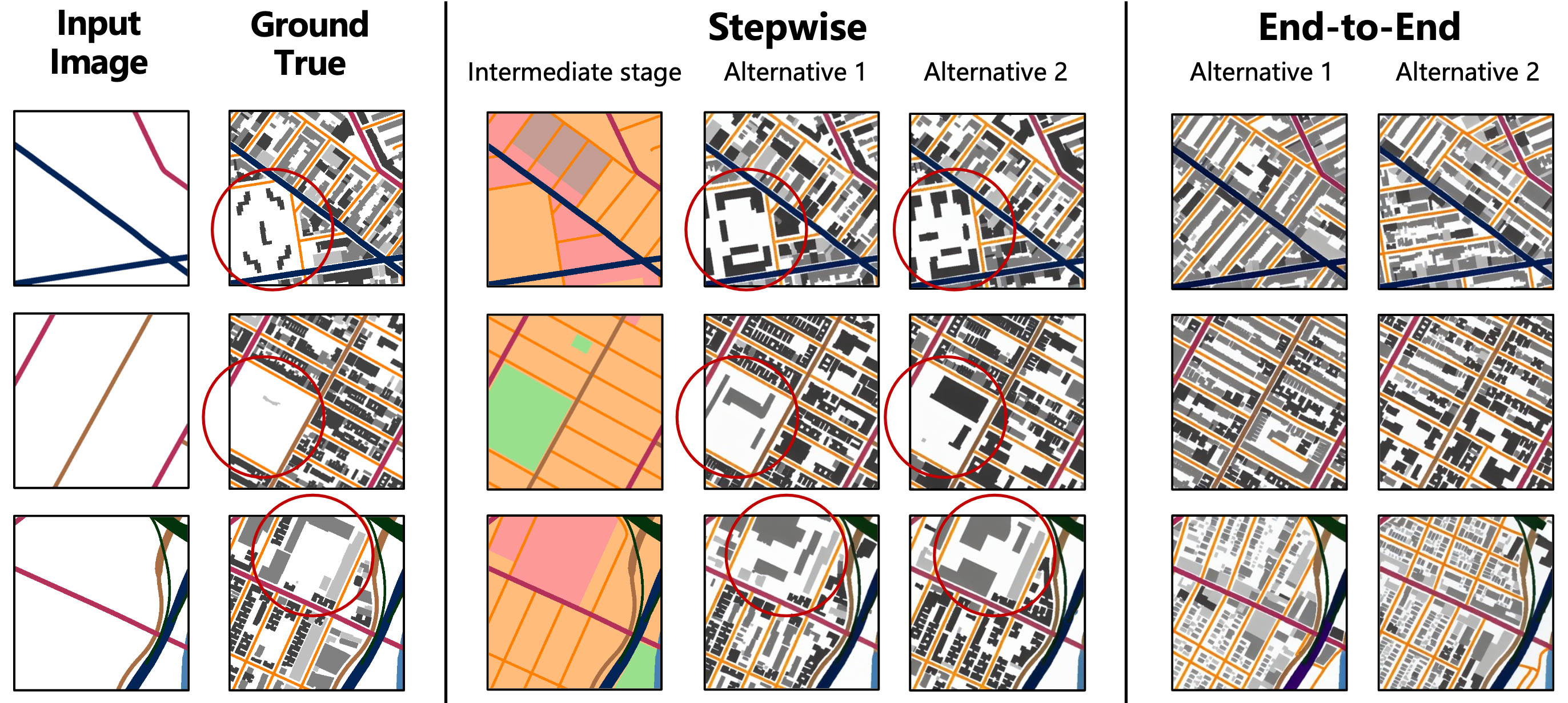}
    \caption{Comparison of generated images using stepwise and end-to-end frameworks}
    \label{fig:res-stepwise}
\end{figure}

Figure~\ref{fig:res-stepwise} compares the visual outputs of the stepwise and end-to-end frameworks. By examining the generated building layouts, it is evident that the stepwise framework demonstrates a clearer understanding of land use assignments, even though both frameworks incorporate land use composition in their prompts. Specifically, the stepwise framework produces distinct layouts for different zones—larger, cohesive buildings for commercial and industrial areas, smaller structures for residential zones, and open areas for parks. In contrast, the end-to-end model, while able to produce reasonable layouts, tends to generate repetitive and overly uniform patterns. This reduces the diversity and functional richness of the generated urban environments. Moreover, the stepwise framework incorporates intermediate stages for generating land use and road network layouts, which allows human designers to review, modify, and refine the outputs. This flexibility is especially valuable in urban design practice, where planners must consider nuanced objectives and context-specific requirements that AI alone may not fully capture.

 \subsection{Design Diversity in Generated Images} \label{sec: res-diversity}

\begin{figure}[!ht]
    \centering
    \includegraphics[width=1.0\linewidth]{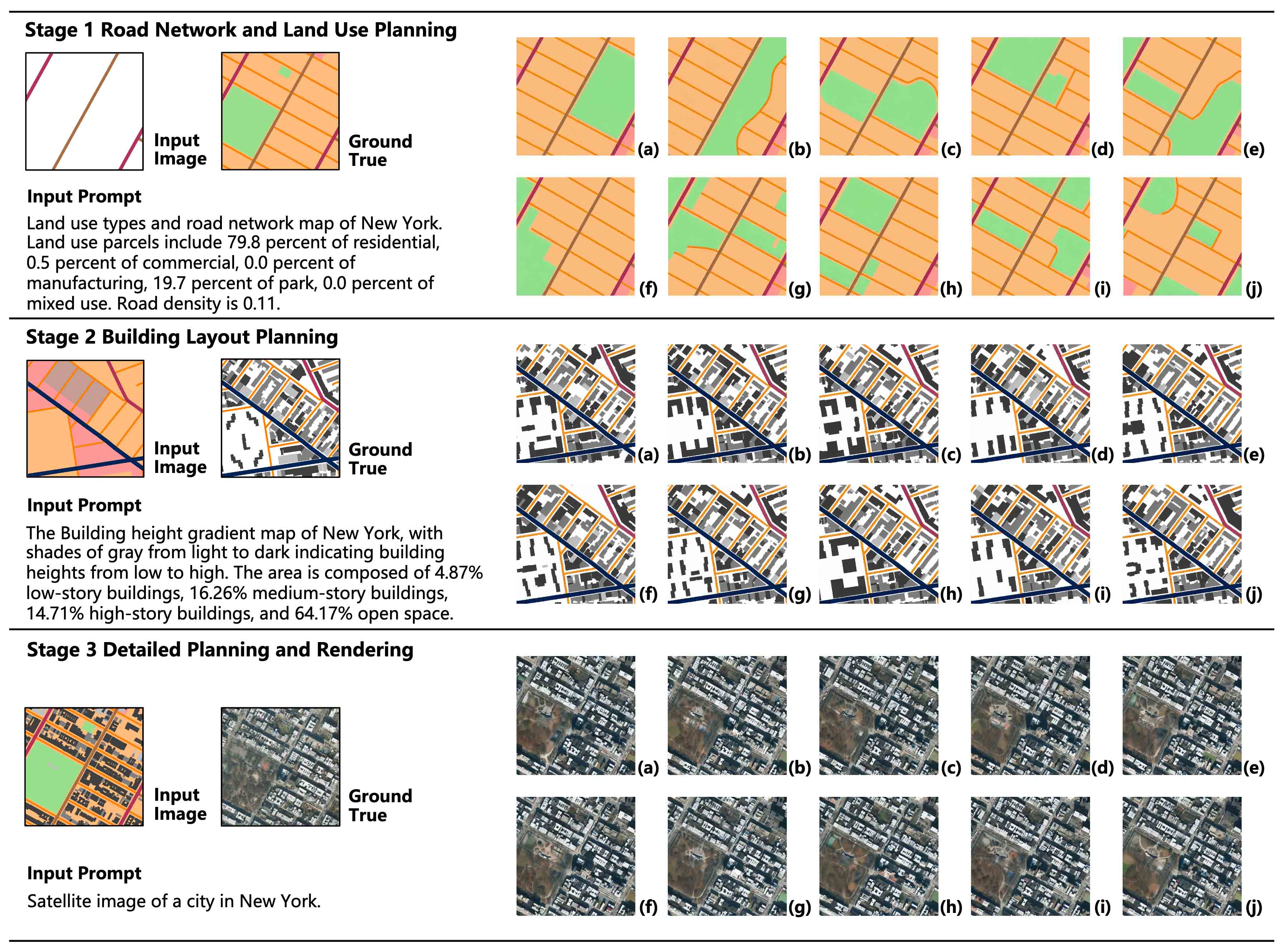}
    \caption{Diversity of Generated Images}
    \label{fig:diversity}
\end{figure}

In this section, we examine ControlNet’s capacity to generate diverse urban designs, highlighting its flexibility and utility for design exploration. As shown in Figure~\ref{fig:diversity}, when provided with the same input images and prompts, ControlNet can produce multiple variations of urban layouts. This allows human designers for comparative selection, thereby efficiently supporting the exploration of design possibilities within the urban design workflow.

In the stage 1 example, the prompt specifies a land use composition of 19.7\% park and 79.8\% residential. While all outputs fulfill the specified instruction, they exhibit a variety of spatial configurations. Alternative (a) features a large, square-shaped central park surrounded by residential areas, closely resembling the layout observed in the ground truth. Alternative (b) adopts a similar centralized configuration, but are aligned with the main road in a linear form, with a curved boundary on the opposite side. Alternatives (c) and (d) divide the park into two segments, maintaining a relatively compact configuration, though with less regular shapes. In contrast, alternatives (g), (h), (i) and (j) explore a more dispersed layout, with smaller individual parks distributed throughout the area. This pattern allows more residential buildings to be situated adjacent to green spaces, offering alternative design possibilities that enhance neighborhood accessibility to open space.

In the stage 2 example, ControlNet explores different building layout patterns for the large, contiguous residential parcel in the lower-left corner of the image. Alternatives (d), (g) and (i) feature linear, row-based arrangements of buildings.  Alternatives (e) and (f) also adopt row-based configurations but introduce larger square-shaped structures at the center, which could serve community functions within the neighborhood. Alternatives (b) and (h) adopt a courtyard-oriented configuration, with "U"-shaped or "L"-shaped building blocks set up along the street, forming a large enclosed or semi-enclosed courtyard in the middle. Alternatives (a), (c), and (j) combine "U"-, "L"-, and linear-shaped buildings to create multiple clusters, with smaller courtyards in between. These varied spatial arrangements offer human designers a rich set of options for organizing residential environments, supporting different functional and aesthetic design goals.

In the stage 3 example, the model is tasked with generating a satellite-style detailed urban design within a block that combines residential and park land uses. The resulting park designs exhibit a wide range of stylistic variations in terms of path layouts, open squares, water features, and green space configurations. For instance, alternatives (a), (c), and (e) feature paths with geometric and symmetrical arrangements, while (f) shows more organic and free-form path patterns. Alternatives (a)-(e) also include central squares designed to accommodate public gatherings, while alternative (h) features a park dominated almost entirely by green space, offering greater exposure to nature. In alternatives (e), (f), and (h), the presence of water bodies of varying sizes further reflects the exploration of diverse possibilities in park design.

These examples also underscore the essential role of human designers in selecting and refining AI-generated design diagrams. In Stage 1, some outputs, such as (g) and (j), produce dead-end roads that may hinder connectivity and reduce transport efficiency. In such cases, human expertise is critical for identifying and resolving functional issues in the design diagram. In Stage 2, the diverse residential arrangements generated by the model offer valuable options, but human designers need to consider building regulations and public feedback to determine the most contextually appropriate solution. In Stage 3, although the satellite-style images showcase the model’s ability to produce detailed urban design elements, further refinement is necessary to clarify the placement and shape of these features and ensure that the output provides actionable, high-quality guidance for implementation.

\subsection{Urban Transferability} \label{sec: res-transfer}

Urban designtransferability holds significant value in urban design. By applying the design style of one city to another, planners can compare urban patterns across different cities to understand their distinct characteristics. It also enables the creation of cross-city design solutions by reimagining one city's site using the planning style of another.

\begin{figure}[!ht]
    \centering
    \includegraphics[width=0.8\linewidth]{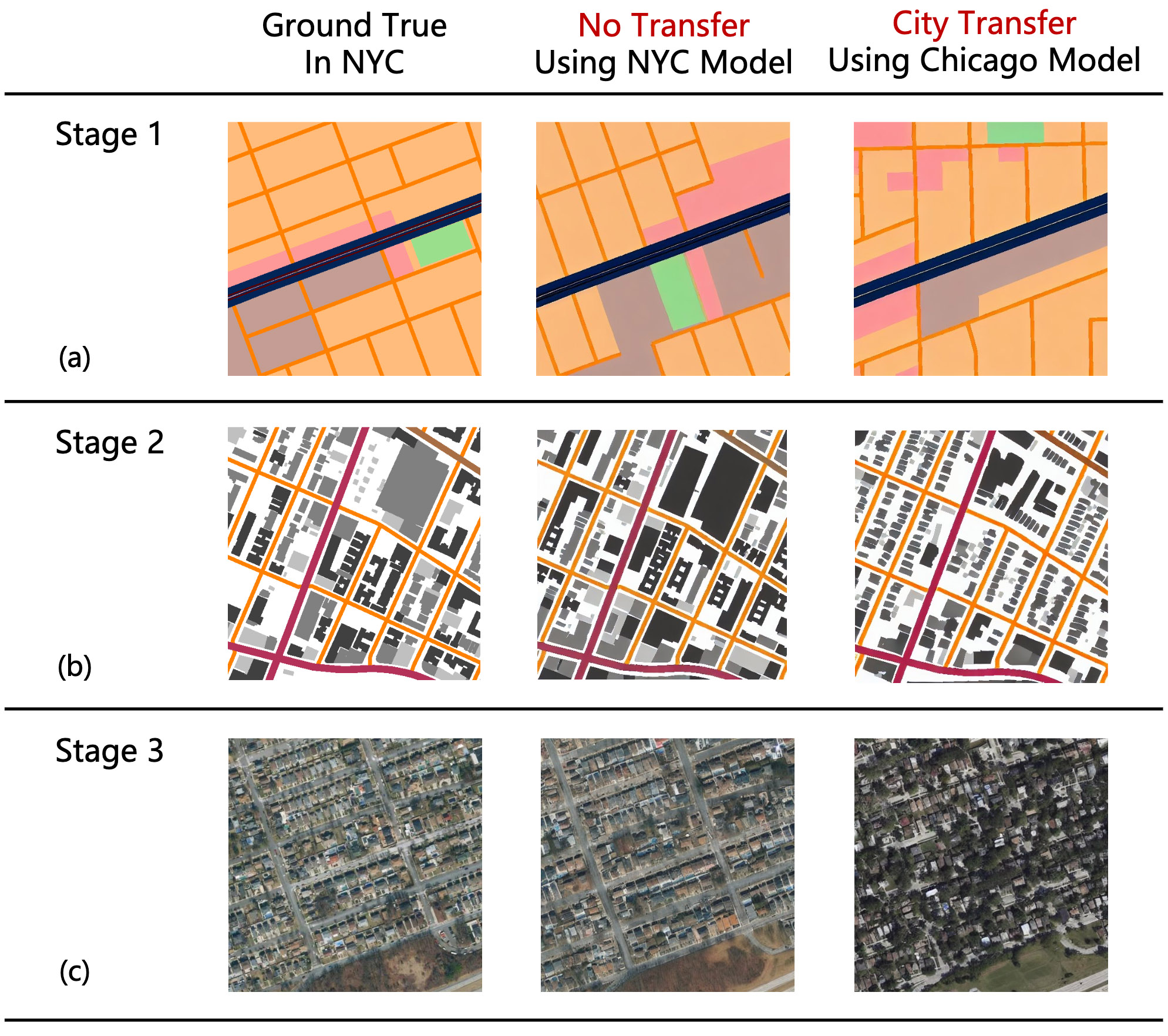}
    \caption{Urban Transferability Results}
    \label{fig:transfer}
\end{figure}

To assess urban design transferability, we apply models trained on Chicago to generate urban design diagrams for sites in NYC, as shown in Figure~\ref{fig:transfer}. The results demonstrate that, under identical site constraints and design instructions, the generated images exhibit distinct urban visual patterns influenced by the design characteristics of the training city.
In the Stage 1 example, the models reflect differences in road network orientation. The NYC model generates obliquely aligned roads, consistent with NYC’s irregular grid. In contrast, the Chicago model produces road networks predominantly aligned along a north-south axis, even when the site constraint includes diagonally aligned major roads, reflecting Chicago’s typical grid pattern. 
In Stage 2, the NYC model generates building footprints with greater variation in size and height, mixing low- and high-rise structures within the same block. In contrast, the Chicago model produces more uniform footprints, reflecting the typical building-height patterns of each city.
In Stage 3, the Chicago model generates more street trees than the NYC model, reflecting the cities' difference in green space arrangements. 
These results illustrate how identical design instructions yield diverse outcomes depending on the urban context, highlighting GenAI’s ability to support cross-city learning by transferring urban design styles from one city to another.

\section{Conclusion and Discussion}
This study introduces a stepwise generative framework for urban design that integrates human expertise with multimodal diffusion models, addressing key limitations of existing GenAI approaches that offer insufficient human control and overlook the iterative and nonlinear nature of urban design. Specifically, we structure the design process into three sequential stages: (1) road network and land use planning, (2) building layout planning, and (3) detailed planning and rendering. At each stage, we adapt the state-of-the-art ControlNet diffusion model to generate urban design layouts guided by site constraints and planning guidance. 
Experimental results from Chicago and New York City demonstrate that the proposed approach outperforms previous GAN-based baselines in terms of visual fidelity, instruction compliance, and visual diversity. Additionally, compared to end-to-end frameworks which generate a final urban design diagram directly from site constraints, our proposed stepwise framework is superior in terms of output quality, alignment with instructions, and better human control. These findings highlight the effectiveness of multimodal diffusion models and stepwise generation in supporting iterative refinement and sustained human control, advancing the integration of GenAI into real-world urban design workflows.

This framework presents several potential applications in urban design and design. 
First, for urban planners, the framework facilitates efficient and diverse generation of alternatives across different design stages. By enabling planners to evaluate and compare options at each stage and feed the refined alternatives into subsequent stages, the framework supports iterative design processes while adhering to human instructions. 
Second, the rapid visualization capability of the proposed framework can enhance public engagement in urban development projects. It enables near real-time visual outputs to support discussions between planners and the public around hypothetical scenarios. For example, how changes in land use might alter urban form, or how variations in building density or height could affect the spatial experience. 
Compared to traditional CAD/BIM tools that require specialized skills, our model provides a more accessible tool for the general public to explore urban design ideas and express their needs. 
Third, the framework facilitates the interpretation of design pattern differences across cities and supports rapid experimentation with urban style transfer. By examining how identical design constraints and objectives yield different outcomes under varying urban contexts, urban designers and policy makers can gain deeper insights into how local conditions shape design solutions. Additionally, the urban style transfer offers an automatic way to learn from other cities by reimagining one city's site using the planning style of another.

While this study advances the integration of generative AI with human expertise in urban design, several limitations remain. First, expanding the dataset to include a wider variety of cities, particularly those with urban structures distinct from the grid patterns commonly found in U.S. cities (e.g., cities in Europe and Asia), would enable the framework to address more diverse regional contexts. 
Second, 
the framework primarily focuses on urban form and does not deeply explore the underlying generative logic of urban spaces, such as the socio-economic, environmental, or cultural factors that fundamentally shape space layout. The current approach emphasizes geometric and visual compliance but is limited in explicit mechanisms to incorporate qualitative, context-driven principles that guide real-world urban space dynamics. Future research can partly address this by integrating more comprehensive and flexible design instructions, including demographic data, socioeconomic indicators, and planning regulations, to provide deeper, context-aware guidance for urban design.
Third, the current framework is more suited to greenfield development than urban renewal. In renewal contexts, design outcomes are heavily influenced by stakeholder dynamics, such as landowner behavior, legal constraints, and fragmented property rights, which introduce unpredictability and complexity beyond urban form. Future work should integrate participatory stakeholder models or negotiation-driven simulations into the generative process, and the updated framework could better adapt to the fluidity of urban renewal, ensuring designs not only visually coherent but also politically and economically actionable.

\appendix
\section{The performance of GPT-4o in generating urban design diagrams}\label{appendix}

\begin{figure}[!ht]
    \centering
    \includegraphics[width=1\linewidth]{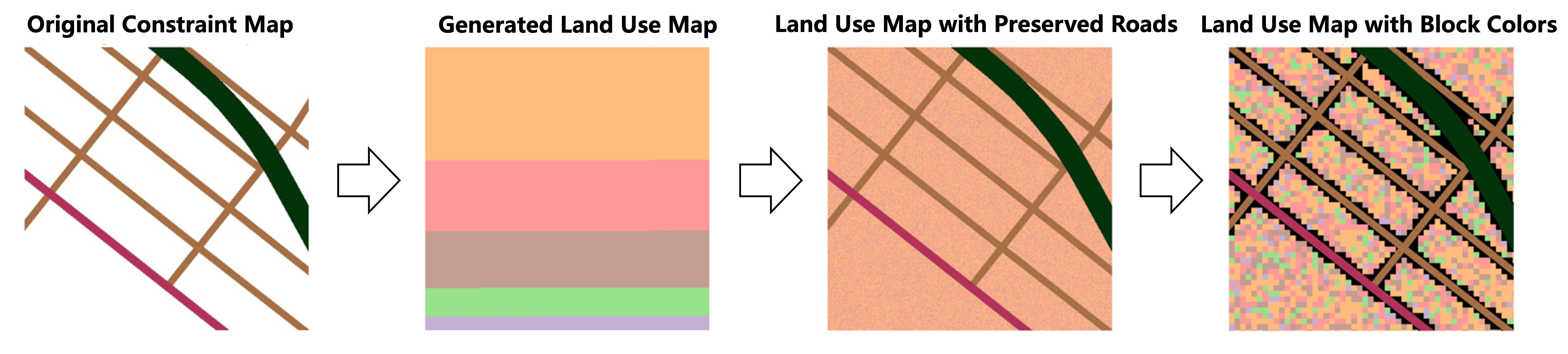}
    \caption{GPT-4o for road network and land use planning}
    \label{fig:gpt}
\end{figure}

This appendix presents the results of using ChatGPT-4o to generate road network and land use maps given site constraint images and textual prompts. As shown in Figure~\ref{fig:gpt}, while ChatGPT-4o demonstrates an ability to interpret land use proportions and RGB values, it produces overly simplified outputs, such as basic color blocks or pixel maps, failing to capture the spatial relationships between roads and land use types. 
This suggests that general-purpose models like ChatGPT-4o may not yet be well-suited for domain-specific scenarios requiring complex spatial reasoning and design coherence.

\section*{Declaration of Generative AI and AI-assisted technologies in the writing process}
During the preparation of this work the authors used ChatGPT in order to polish the grammar and wording of part of the manuscript. After using this tool/service, the authors reviewed and edited the content as needed and take full responsibility for the content of the publication.

\section*{Acknowledgements}
We thank Singapore-MIT Alliance for Research and Technology (SMART) for funding this research, and thank Prof. Zhan Zhao for partially supporting our computing resources.






\bibliographystyle{model5-names2}
\biboptions{authoryear}
\bibliography{main}







\end{document}